\documentclass[10pt,twocolumn,letterpaper]{article}

\pdfoutput=1

\usepackage{iccv}
\usepackage{times}
\usepackage{epsfig}
\usepackage{graphicx}
\usepackage{amsmath}
\usepackage{amssymb}

\usepackage{tabularx}
\usepackage{makecell}
\usepackage{multirow}
\usepackage{subfig}
\usepackage{tabulary}
\usepackage{booktabs}

\usepackage[breaklinks=true,bookmarks=false]{hyperref}

\iccvfinalcopy 

\ificcvfinal\fi

\begin{document}

\title{Team PFDet's Methods for Open Images Challenge 2019}

\author{
Yusuke Niitani$^*$\,\,\,
Toru Ogawa$^*$\,\,\,
Shuji Suzuki\thanks{The starred authors contributed equally and are ordered alphabetically.}\\
Takuya Akiba\,\,\,
Tommi Kerola\,\,\,
Kohei Ozaki\,\,\,
Shotaro Sano\\
Preferred Networks, Inc.\\
{\tt\small \{niitani,ogawa,ssuzuki,akiba,tommi,ozaki,sano\}@preferred.jp}
}

\maketitle
\ificcvfinal\thispagestyle{empty}\fi

\begin{abstract}
We present the instance segmentation and the object detection method used by team PFDet for Open Images Challenge 2019.
We tackle a massive dataset size, huge class imbalance and federated annotations.
Using this method, the team PFDet achieved 3rd and 4th place in the instance segmentation and the object detection track, respectively. 
\end{abstract}

\section{Introduction}
Open Images Detection Dataset V5 (OID)~\cite{kuznetsova2018openimages} is currently the largest publicly available object detection dataset, including $1.7$M annotated images with $12$M bounding boxes.
The diversity of images in training datasets is the driving force of the generalizability of machine learning models. Successfully trained models on OID would push the frontier of object detectors with the help of data.

Since the number of images in OID is extremely large, the speed of training is critical.
For faster training, we use Fast R-CNN~\cite{girshick2015fast} instead of the more commonly used Faster R-CNN~\cite{ren2015faster}.
Fast R-CNN omits time-consuming online RoI (Region of Interest) generation during training by pre-computing RoIs.
We find that the selection of pre-computed RoIs plays an important role in achieving good accuracy.
For instance, when the number of pre-computed RoIs is small during training, a network overfits to those RoIs.
By default, RoIs used to train Faster R-CNN have high variation, so the aforementioned problem is unique to Fast R-CNN.

OID is a federated object detection dataset~\cite{gupta2019lvis, kuznetsova2018openimages}.
This means that for each image, only a subset of categories is annotated.
This is in contrast to exhaustively annotated datasets such as COCO~\cite{yi2014coco}.
Federated annotation is a realistic approach to expand the number of categories covered by the dataset, since without sparsifying the number of annotated categories, the number of annotations required may explode as the total number of categories increases.

When training a detector on exhaustively annotated datasets like COCO, the loss functions make an assumption that no objects is inside an unannotated region.
However, for federated object detection datasets, such assumption may be violated because some regions contain an object of an unverified category.
We handle this problem by ignoring loss for unverified categories~\cite{niitani2019sample}.

In addition to the previously mentioned uniqueness of OID, the dataset poses an unprecedented class imbalance for an object detection dataset.
The rarest category \textit{Pressure Cooker} is annotated in only $13$ images, but the most common category \textit{Person} is annotated in more than $800$k images.
The ratio of the occurrence of the most common and the least common category is $183$ times larger than in COCO~\cite{yi2014coco}.
Typically, this class imbalance can be handled by by over-sampling images containing rare categories.
However, over-sampling may suffer from degraded performance for common categories.

As a practical method to solve class imbalance, we train models exclusively on rare classes and ensemble them with the rest of the models similar to our our method in the last year's competition~\cite{akiba2018pfdet}.

To summarize our major contributions:
\begin{itemize}
\item \textbf{Fast R-CNN}: We present the effectiveness of Fast R-CNN and propose methods to alleviate performance penalty introduced by per-computed RoIs.
\item \textbf{Using Only Verified Categories}: We find that it is helpful to ignore unverified categories during training.
\item \textbf{Expert Models}: We show the effectiveness of using expert models, especially for classes that rarely appear in the dataset.
\end{itemize}

\section{Methods}

\subsection{Model architecture}
Two stage object detectors such as Faster R-CNN~\cite{ren2015faster} are known to achieve excellent accuracy, but their GPU usage efficiency during training is sub-optimal.
This is because RoIs used to train R-CNN heads are determined between feature extraction and loss calculation. Thus, GPUs are forced to wait for the ground-truth assignment of RoIs.
To make a more efficient use of GPUs, we use Fast R-CNN, which pre-computes RoIs and assigns ground-truths to RoIs in parallel to feature extraction.

For the instance segmentation track, we add a mask head~\cite{he2017maskrcnn} that predicts a segmentation of an object given a region around it and its category.
The $300$ categories for instance segmentation is a subset of the $500$ categories for detection.
We use all the $500$ detection categories even for training the instance segmentation model, where only $300$ of these have instance masks.
This worked better compared to only using the $300$ instance segmentation categories in preliminary experiments.

\subsection{Learning with a federated dataset}
In federated object detection datasets~\cite{kuznetsova2018openimages,gupta2019lvis}, for each image, categories are grouped into positively verified, negatively verified and unverified.
For positively verified categories, annotations are exhaustively made for all objects of those categories.
For negatively verified categories, the annotators have made sure that the image contains no objects of those categories.
For unverified categories, the objects of those categories may or may not exist in the image.

During training, each RoI is assigned to one of the ground truth boxes if there is any that has sufficiently large enough intersection.
This assignment is used to calculate classification loss and localization loss~\cite{ren2015faster}. In this work, the classification loss is calculated as the sum of sigmoid cross entropy loss for each proposal and each category $c$ as:
\begin{equation}
\begin{aligned}
\mathcal{L}_{cls} &= - \sum_i \sum_c l_{ic}\log p_{ic} \\
l_{ic} &\in \{-1, 0, 1\}~,
\end{aligned}
\end{equation}
where $l_{ic} = 1$ and $l_{ic} = -1$ when the $i$-th RoI is assigned or not assigned to the category $c$, respectively.
Also, $l_{ic}$ can be set to $0$, which means that the classification loss for the category $c$ is ignored for the $i$-th RoI.

When the RoI $i$ is not assigned to any of the ground truth boxes, we set $l_{ic} = -1$ for positively and negatively verified categories and $l_{ic} = 0$  for unverified categories.
When the RoI $i$ is assigned to a ground truth bounding box with the category $c'$, we set $l_{ic'} = 1$ and $l_{ic} = -1$ for all negatively verified and positively categories except for the category $c'$.
We set $l_{ic} = 0$ for unverified categories.
In practice, verified categories are expanded based on the category hierarchy.

\subsection{Expert models}
In OID, there is an extreme class imbalance, which makes it difficult for a model to learn rare categories.
For instance, there are $238$ classes that are annotated in less than $1000$ images, but the most common class \textit{Person} is annotated in $807$k images.
We use expert models fine-tuned from a model trained with the entire dataset as done in our previous year's submission~\cite{akiba2018pfdet}.

We select a subset of categories to which an expert model is trained based on one of the following criteria:
\begin{itemize}
    \item Occurrence ranking of categories in the detection subset. We group categories that are in a neighboring ranking so that sampling imbalance does not occur among the categories in a subset.
    \item Occurrence ranking in the instance segmentation subset.
    \item Semantic similarity of categories. We cluster categories based on their similarity of embeddings by an imagenet pretrained feature extractor~\cite{ouyan2016expert}.
\end{itemize}

When training an expert model, the annotations of categories that the expert is not responsible for are dropped.
Images that do not contain an annotation of target categories are discarded.
Also, the bias term of the classification layer is reinitialized so that the network only outputs target categories.

\subsection{Ensembling}
When aggregating predictions from multiple models, we apply non-maximum suppression to predictions from each model and apply suppression once again to concatenation of the predictions.
In the second suppression step, we group predictions that have a large enough intersection and are assigned to the same category.
We compute a representative prediction from each group, and the collection of the representatives is the final prediction.
Given a group, the bounding box of the group's representative is the bounding box of the most confident prediction in the group.
The segmentation is calculated as the average of the segmentations in the group weighted by confidence and spatial proximity.

\subsection{Pre-computed RoIs}
The RoIs used by Faster R-CNN are conditioned on the model weights.
Thus, high variation of RoIs could be achieved without any effort.
This variation is lost when we use Fast R-CNN models with the same set of pre-computed RoIs.
This comes at the cost of degraded performance.
During training training, RoIs that are used to compute head losses are sampled from a pool of RoIs.
For training Fast R-CNN, the pool of RoIs is pre-computed, and we find that the number of RoIs in the pool needs to be very high in order for the network not to overfit to the RoIs.
In many published works using a variant of Faster R-CNN~\cite{he2017maskrcnn}, the number of RoIs in the pool for each image is up to $2000$.
However, we find that this is not large enough, so we prepare up to $16000$ RoIs per image.

Selection of pre-computed RoIs is also important when ensembling models.
The set of RoIs used by each model should be different when ensembling.
This is because predictions made with different set of RoIs complement each other better.

\subsection{Post-processing}
As stated in the competition description page, the size of an annotated object is larger than $40\times80$ or $80\times40$~\footnote{\url{https://storage.googleapis.com/openimages/web/factsfigures.html}}
Thus, any predictions with small segmentations are unlikely to be counted as true positives during evaluation.
We omitted predictions whose areas of their segmentation are less than $1600$ pixels.

The competition submission file size is limited to $5$GB.
Our submission file sometimes exceeded this limit, especially after ensembling.
We find that predictions for some categories occur much more frequently than the rest, which are likely to be less important for a class averaged metric.
We drop predictions of frequently predicted categories to meet the file size limit.

\section{Experiments}
We used COCO~\cite{yi2014coco}, LVIS~\cite{gupta2019lvis} and Objects365 as the external data.
We use Feature Pyramid Networks~\cite{lin2017feature} for our experiments.
The feature extractor is SENet~\cite{hu2017squeeze}.
The initial bias for the final classification layer is set to a large negative number to prevent unstable training in the beginning.
We set the initial weight of the base extractor by the weights of an image classification network trained on the ImageNet classification task~\cite{imagenet_cvpr09}.
We use stochastic gradient descent with the momentum set to $0.9$ for optimization.
The base learning rate is set to $0.00125 \times \text{batchsize}$.
We used up to $120$ GPUs. The best single model is trained with the batch size set to $240$.
We used multi-node batch normalization~\cite{megdet} to make training stable.
We trained for $16$ epochs.
The learning rate is scheduled by a cosine function $\eta = \eta_0 \frac{\cos{(\text{\% of progress} \times \pi)} + 1}{2}$, where $\eta$ and $\eta_0$ are the learning rate and the initial learning rate.
We scale images during training so that the length of the smaller edge is between $[650, 1056]$.
Also, we randomly flip images horizontally to augment training data. In addition to that, for training expert models, we used an augmentation policy searched by AutoAugment~\cite{zoph2019autoaug}.
During inference, we did not do any test-time augmentation.
We used non-maximum suppression with the intersection over union threshold set to $0.5$.
We use Chainer~\cite{tokui2019chainer,akiba2017chainermn,niitani2017chainercv} in our experiments.

\subsection{Pre-computed RoIs}
To study the best ensembling strategy of Fast R-CNN models, we conducted an ablative study of ensembles of  predictions from two models with two sets of RoIs.
The result is shown in Table~\ref{tab:ensemble}.
When ensembling predictions from two models using the same set of RoIs, the performance did not improve from the single model result.
By using different sets of RoIs, the ensemble outperformed the single model.
\begin{table}
\centering \addtolength{\tabcolsep}{-2pt}
\footnotesize
\caption{Comparison of ensembling results with different set of RoIs. The top three rows show the scores with single model with different weights and RoIs. The bottom two rows show the scores of ensembling predictions from two models. The last row shows the result when different set of RoIs are used for predictions of two models.
}\label{tab:ensemble}
\begin{tabular}{lcc}
\toprule
Model  & RoI & Segmentation val mAP  \\
\midrule
 A  &  1 &  $70.62$ \\
 B  &  1 &  $71.03$ \\
 B  &  2 &  $71.02$ \\
 A, B  & 1, 1 &  $71.02$ \\
 A, B  & 1, 2 &  $\mathbf{71.78}$ \\
\bottomrule
\end{tabular}\vspace{0.1cm}
\vspace{-0.03in}
\end{table}

\subsection{Expert models}
Table~\ref{tab:experts} shows an ablative study of expert models with different numbers of categories assigned.
This is the result when training expert models for the 50th to 99th rarest categories.
When training multiple expert models for these categories, the categories are split into disjoint sets. These splits are made based on how frequent the categories appear in OID.
For instance, when training two expert models, the first expert model is responsible for the 50th to 74th rarest categories and the second model is responsible for the 75th to 99th rarest categories.
As seen in the table, when each expert is responsible for a smaller number of categories, the performance for each category improves on average.
Since the computational budget is limited, it is difficult to make the number of categories assigned to each expert small.
Thus, the numbers of categories responsible by an expert model in our final submission vary.
\begin{table}
\centering \addtolength{\tabcolsep}{-2pt}
\footnotesize
\caption{Ablative study on the number of categories assigned to expert models. 
The mean average precision of the baseline model is $65.39$.
}\label{tab:experts}
\begin{tabular}{ccc}
\toprule
\# of categories per expert   & \# of experts & detection val mAP  \\
\midrule
 50  &  1 &  $69.19$ \\
 25  &  2 &  $70.45$ \\
 10  &  5 &  $71.37$ \\
\bottomrule
\end{tabular}\vspace{0.1cm}
\vspace{-0.03in}
\end{table}

\subsection{Competition results}
Our final submission consists of the predictions from the following models:
\begin{itemize}
    \item Two Fast R-CNN models trained on $500$ detection categories. One of them is trained for $16$ epochs and another is trained for $24$ epochs.
    \item $47$ Fast R-CNN expert models. On average each expert predicts $43$ categories.
    \item Faster R-CNN models. We used predictions from Faster R-CNN models trained in the preliminary experiments. Some of them are from the last year's submission~\cite{niitani2019sample,akiba2018pfdet}.
\end{itemize}

\vspace{-0.15cm}
The results for instance segmentation and object detection are shown in Table~\ref{tab:segmentation} and Table~\ref{tab:detection}.
We ranked 3rd and 4th place in the instance segmentation and the object detection tracks, respectively.

For instance segmentation, all ensemble results exceeded the file size limit of 5GB for the test set. Thus, we needed to drop some predictions for the frequently predicted categories.
Therefore, by adding more models, the instance segmentation test results did not improve as much as the validation scores and the object detection results.

\begin{table}
\centering \addtolength{\tabcolsep}{-2pt}
\footnotesize
\caption{Mean average precision on the instance segmentation track.
We did not set a file size limit for the validation set, so the post processing of removing small masks was not evaluated on the validation set.
Some of the predictions of Faster R-CNN models are from last year's competition, so we could not evaluate them on the validation set.
}\label{tab:segmentation}
\vspace{-0.20cm}
\begin{tabular}{l|cccc}
\toprule
 & val & public test & private test  \\
\midrule
Full (16 epochs)  &  $70.62$ & $51.33$ & $46.33$ \\
Full (24 epochs)  &  $71.02$ & $51.80$ & $47.17$ \\
Ensemble of above two &  $71.61$ & $52.67$ & $47.32$ \\
+ Expert Models &  $75.74$ &  $54.55$ & $50.55$ \\
+ Remove small masks &   &  $54.83$ & $50.76$ \\
+ Faster R-CNN models &   & $\mathbf{55.33}$ & $\mathbf{51.10}$ \\ 
\bottomrule
\end{tabular}
\vspace{-0.03in}
\end{table}
\begin{table}
\centering \addtolength{\tabcolsep}{-2pt}
\footnotesize
\caption{Mean average precision on the detection track.
Some of the predictions of Faster R-CNN models are from last year's competition, so we could not evaluate them on the validation set.
}\label{tab:detection}
\begin{tabular}{l|cccc}
\toprule
 & val & public test & private test  \\
\midrule
Full (16 epochs)  &  $68.05$ & $59.03$ & $55.81$ \\
Full (24 epochs)  &  $68.44$ & $59.32$ & $56.32$ \\
Ensemble of above two &  $69.02$ & $60.31$ & $57.14$ \\
+ Expert Models  & $73.10$ & $64.54$ & $61.26$ \\
+ Faster R-CNN models &   & $\mathbf{65.45}$ & $\mathbf{62.22}$ \\ 
\bottomrule
\end{tabular}\vspace{0.1cm}
\vspace{-0.2in}
\end{table}

\vspace{-0.1in}
\section{Conclusion}
In this paper, we described the instance segmentation and the object detection submissions to Open Images Challenge 2019 by th e team PFDet.
Thanks to the fast research cycle enabled by an efficient usage of large GPU clusters,
we developed several techniques that led to 3rd and 4th place in the the instance segmentation and the object detection track, respectively.

\vspace{-0.25cm}
{
\small
\paragraph{Acknowledgments}
We thank K.~Uenishi, R.~Arai, T.~Shiota and S.~Omura for helping with our experiments.
}
\vspace{-0.15cm}

{\small
\bibliographystyle{ieee_fullname}
\bibliography{egbib}

\begin{thebibliography}{10}\itemsep=-1pt

\bibitem{akiba2017chainermn}
Takuya Akiba, Keisuke Fukuda, and Shuji Suzuki.
\newblock {ChainerMN: Scalable Distributed Deep Learning Framework}.
\newblock In {\em LearningSys workshop in NIPS}, 2017.

\bibitem{akiba2018pfdet}
Takuya Akiba, Tommi Kerola, Yusuke Niitani, Toru Ogawa, Shotaro Sano, and Shuji
  Suzuki.
\newblock Pfdet: 2nd place solution to open images challenge 2018 object
  detection track.
\newblock In {\em ECCV Workshop}, 2018.

\bibitem{zoph2019autoaug}
Golnaz Ghiasi Tsung-Yi Lin Jonathon Shlens Quoc V.~Le Barret~Zoph, Ekin
  D.~Cubuk.
\newblock Learning data augmentation strategies for object detection.
\newblock In {\em arxiv}, 2019.

\bibitem{imagenet_cvpr09}
J. Deng, W. Dong, R. Socher, L.-J. Li, K. Li, and L. Fei-Fei.
\newblock {ImageNet: A Large-Scale Hierarchical Image Database}.
\newblock In {\em CVPR}, 2009.

\bibitem{girshick2015fast}
Ross Girshick.
\newblock Fast r-cnn.
\newblock In {\em ICCV}, 2015.

\bibitem{gupta2019lvis}
Agrim Gupta, Piotr Dollár, and Ross Girshick.
\newblock Lvis: A dataset for large vocabulary instance segmentation.
\newblock In {\em CVPR}, 2019.

\bibitem{he2017maskrcnn}
Kaiming He, Georgia Gkioxari, Piotr Dollár, and Ross Girshick.
\newblock Mask r-cnn.
\newblock In {\em ICCV}, 2017.

\bibitem{hu2017squeeze}
Jie Hu, Li Shen, and Gang Sun.
\newblock Squeeze-and-excitation networks.
\newblock {\em CVPR}, 2018.

\bibitem{kuznetsova2018openimages}
Alina Kuznetsova, Hassan Rom, Neil Alldrin, Jasper Uijlings, Ivan Krasin, Jordi
  Pont-Tuset, Shahab Kamali, Stefan Popov, Matteo Malloci, Tom Duerig, and
  Vittorio Ferrari.
\newblock The open images dataset v4: Unified image classification, object
  detection, and visual relationship detection at scale.
\newblock {\em arXiv:1811.00982}, 2018.

\bibitem{lin2017feature}
Tsung-Yi Lin, Piotr Doll{\'a}r, Ross~B Girshick, Kaiming He, Bharath Hariharan,
  and Serge~J Belongie.
\newblock Feature pyramid networks for object detection.
\newblock In {\em CVPR}, 2017.

\bibitem{yi2014coco}
Tsung-Yi Lin, Michael Maire, Serge Belongie, Lubomir Bourdev, Ross Girshick,
  James Hays, Pietro Perona, Deva Ramanan, C.~Lawrence Zitnick, and Piotr
  Dollár.
\newblock Microsoft coco: Common objects in context.
\newblock {\em ECCV}, 2014.

\bibitem{niitani2019sample}
Yusuke Niitani, Takuya Akiba, Tommi Kerola, Toru Ogawa, Shotaro Sano, and Shuji
  Suzuki.
\newblock Sampling techniques for large-scale object detection from sparsely
  annotated objects.
\newblock In {\em CVPR}, 2019.

\bibitem{niitani2017chainercv}
Yusuke Niitani, Toru Ogawa, Shunta Saito, and Masaki Saito.
\newblock Chainercv: a library for deep learning in computer vision.
\newblock In {\em ACM MM}, 2017.

\bibitem{ouyan2016expert}
Wanli Ouyang, Xiaogang Wang, Cong Zhang, and Xiaokang Yang.
\newblock Factors in finetuning deep model for object detection.
\newblock In {\em CVPR}, 2016.

\bibitem{megdet}
Chao Peng, Tete Xiao, Zeming Li, Yuning Jiang, Xiangyu Zhang, Kai Jia, Gang Yu,
  and Jian Sun.
\newblock Megdet: A large mini-batch object detector.
\newblock In {\em CVPR}, 2018.

\bibitem{ren2015faster}
Shaoqing Ren, Kaiming He, Ross Girshick, and Jian Sun.
\newblock Faster r-cnn: Towards real-time object detection with region proposal
  networks.
\newblock In {\em NIPS}, 2015.

\bibitem{tokui2019chainer}
Seiya Tokui, Ryosuke Okuta, Takuya Akiba, Yusuke Niitani, Toru Ogawa, Shunta
  Saito, Shuji Suzuki, Kota Uenishi, Brian Vogel, and Hiroyuki
  Yamazaki~Vincent.
\newblock Chainer: A deep learning framework for accelerating the research
  cycle.
\newblock In {\em KDD}, 2019.

\end{thebibliography}
}

\end{document}